\documentclass[letterpaper, 10 pt, conference]{IEEEtran}  % Comment this line out if you need a4paper

\IEEEoverridecommandlockouts                              % This command is only needed if 
\usepackage{cite}
\usepackage{comment}
\usepackage{amsmath,amssymb,amsfonts}
\usepackage[ruled]{algorithm2e}
\usepackage{algorithmic}
\usepackage{graphicx}
\usepackage{textcomp}
\usepackage{float}
\def\BibTeX{{\rm B\kern-.05em{\sc i\kern-.025em b}\kern-.08em
    T\kern-.1667em\lower.7ex\hbox{E}\kern-.125emX}}
\usepackage{caption} 
\title{\LARGE \bf
Learned Block Iterative Shrinkage Thresholding Algorithm for Photothermal Super Resolution Imaging
}

\author{Samim Ahmadi$^{1}$,\thanks{$^{1}$Samim Ahmadi, Jan Christian Hauffen and Mathias Ziegler are with the Bundesanstalt f{\"u}r Materialforschung und -pr{\"u}fung, Berlin, Germany {\tt\small samim.ahmadi@bam.de}} Jan Christian Hauffen$^{1}$, Linh K{\"a}stner$^{2}$\thanks{$^{2}$Linh K{\"a}stner is with the Chair Industry Grade Networks and Clouds, Faculty of Electrical Engineering, and Computer Science,	
Berlin Institute of Technology, Berlin, Germany}, Peter Jung$^{3}$, \thanks{$^{3}$Peter Jung and Giuseppe Caire are with the Chair Communication and Information Theory, Berlin Institute of Technology, Berlin, Germany}Giuseppe Caire$^{3}$, Mathias Ziegler$^{1}$
}

\begin{document}

\maketitle
\thispagestyle{empty}
\pagestyle{empty}

% !TeX encoding = utf-8
% !TeX language = en_GB
% !TeX spellcheck = en_GB
% !TeX root = paper.tex

\begin{abstract}

Block-sparse regularization is already well-known in active thermal imaging and is used for multiple measurement based inverse problems. The main bottleneck of this method is the choice of regularization parameters which differs for each experiment. To avoid time-consuming manually selected regularization parameter, we propose a learned block-sparse optimization approach using an iterative algorithm unfolded into a deep neural network. More precisely, we show the benefits of using a learned block iterative shrinkage thresholding algorithm that is able to learn the choice of regularization parameters. In addition, this algorithm enables the determination of a suitable weight matrix to solve the underlying inverse problem. Therefore, in this paper we present the algorithm and compare it with state of the art block iterative shrinkage thresholding using synthetically generated test data and experimental test data from active thermography for defect reconstruction. Our results show that the use of the learned block-sparse optimization approach provides smaller normalized mean square errors for a small fixed number of iterations than without learning. Thus, this new approach allows to improve the convergence speed and only needs a few iterations to generate accurate defect reconstruction in photothermal super resolution imaging.

\end{abstract}
% !TeX encoding = utf-8
% !TeX language = en_GB
% !TeX spellcheck = en_GB
% !TeX root = paper.tex

\section{INTRODUCTION}
\label{sec:introduction}
Sparse signal reconstruction is becoming increasingly popular, especially in the industrial sector. These algorithms are highly attractive as in the field of compressed sensing knowledge of the device or target can be exploited for a high-quality reconstruction \cite{cs,cs2}. Consequently, in the last few years optimization algorithms like ISTA, FISTA \cite{fista}, learned algorithms such as LISTA \cite{lista} or adaptions like ALISTA \cite{alista} or LISTA-AT \cite{lista-at} gained much attention, especially in the field of computer vision.
Active thermal imaging for defect detection in the field of nondestructive testing has been one area of application of block optimization in the recent years \cite{murray,APL,qirt,ndte,ole,nsr}. In active thermal imaging, a specimen with defects, such as voids or cracks, is investigated by actively heating up the specimen (e.g. by illuminating it using lasers or flash lamps) and observing the heat diffusion process with an infrared (IR) camera. The observed heat flow by the IR camera is then evaluated using the generated thermal film sequence. Defect indications are clearly visible in these thermal film sequences if the heat accumulates at a void or if changes in the heat flow gradient due to a crack are visible. Knowing the resulting heat distribution by active thermal imaging for a defect-free body, the exact defect distribution of the defective body can be found by solving an underlying inverse problem. This severely ill-posed thermal inverse problem can be solved by common techniques such as singular value decomposition or Bayesian approaches \cite{inverse_prob,bayes}. However, these techniques are computationally intensive and suffer from prediction inaccuracy for small calculation times in contrast to sparsity-exploiting techniques e.g. least absolute shrinkage and selection operator (LASSO) based methods. Especially for multiple-measurement-vector (MMV) problems, group LASSO as a block regularization method can be applied to find an accurate solution for the underlying inverse problems \cite{group_lasso,eldar}. 
Using multiple and different blind illumination patterns for heating in active thermal imaging results in an MMV problem. Blind illumination patterns reflect the assumption that in industrial applications the exact position of illumination is not known \cite{qirt}. This occurs, for example, when the heat source or the specimen is held by a robot that has a certain amount of positional noise. 

Block regularization showed fascinating results in the near past in photothermal super resolution imaging \cite{ndte,ole,nsr} where the goal is to separate closely neighboring features with a spatial resolution significantly better than achievable with a single measurement problem. Unfortunately, the choice of the regularization parameters was made by hand and has to be done very carefully due to the sensitivity of the thresholding \cite{ole}. Further, the algorithm needs hundreds of iterations to reach convergence even if the right regularization parameters have been found.

In this work, a solution is provided for the above mentioned problem by taking a learned iterative joint sparsity approach called LBISTA (learned block iterative shrinkage thresholding algorithm). This approach is realized by applying neural networks to learn e.g. a suitable choice for the regularization parameters or the training weights. The training data is generated synthetically using uniformly random distributed defect distribution and corresponding thermal film sequences. Thus, the use of LBISTA enables a higher convergence speed than the so far used block iterative shrinkage thresholding algorithm (BISTA) without learning.

\section{Mathematical model in active thermal imaging}
\begin{figure}[h]
    \centering
    \includegraphics{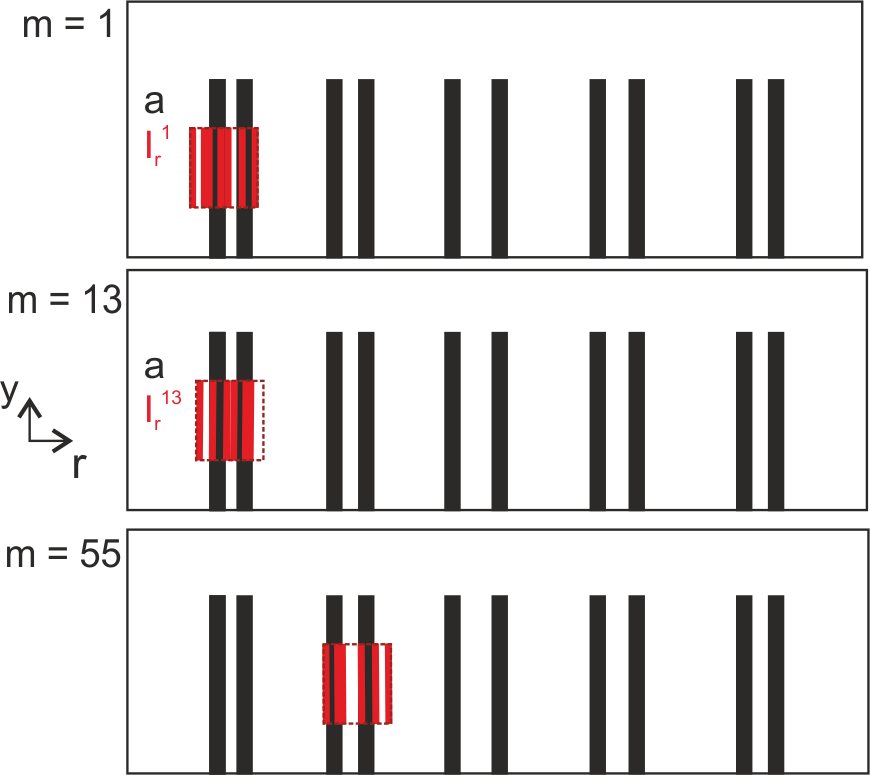}
    \caption{Exemplary specimen with defects shown as blackened stripes. A laser line array with twelve laser lines is used. The illumination pattern differs from measurement to measurement in the illuminated position and the number of laser lines (indicated by red area) which are switched on (randomly chosen). The dashed frame around the pattern indicates the covered illuminated area if all laser lines would be switched on.}
    \label{specimen}
\end{figure}
In these studies, we work with a defect pattern which does not change over the height (lines as defects, see Fig. \ref{specimen}) so that we can simplify our model by calculating the mean over the vertically arranged pixels (see dimension y in Fig. 1 is therefore not considered in the following). The IR camera measures temperature values that can be described in active thermal imaging by a convolution in space and time of the fundamental solution of the heat diffusion equation and the absorbed excitation energy. In the following these measured temperature values are described by discrete matrices: 
\begin{equation}\label{T}
    T = \Phi_{\text{PSF}} \ast_{r,t} x.
\end{equation}
${\Phi_{\text{PSF}} \in \mathbb{R}^{N_r \times N_t}}$ represents the discrete equivalent of the fundamental solution of the heat diffusion equation, where $N_r$ stands for the number of measured pixels in the dimension $r$ and $N_t$ for the number of measured time frames. PSF indicates that the values in the matrix refer to the well-known thermal point spread function (PSF) that can be described mathematically by a Green's function which is known analytically \cite{psf}. The discrete equivalent of the absorbed heat flux density ${x \in \mathbb{R}^{N_r \times N_t}}$ considers the irradiance in discrete space (r) and time (t) dimension ${I_{r,t} \in \mathbb{R}^{N_r \times N_t}}$ as well as the absorption coefficient of the material under investigation in discrete space ${a \in \mathbb{R}^{N_r}}$ by a Hadamard-product: ${x = I_{r,t} \circ a}$.

As described in the introduction, we are dealing with blind structured illumination which leads to the fact that we do not know the exact illuminated positions at the specimen under investigation using a laser as heat source. This results in a small but relevant change of equation \eqref{T} into: 
\begin{equation} \label{eq a}
    T^m = \Phi_{\text{PSF}} \ast_{r,t} x^m = \Phi_{\text{PSF}} \ast_{r,t} (I^m_{r,t} \circ a),
\end{equation}
with ${m = {1 \dots N_{\text{meas}}}}$, whereby $m$ indicates the measurement number and $N_{\text{meas}}$ the number of measurements. The spatial and temporal distribution of the absorbed heat flux density ${x^m = I^m_{r,t} \circ a}$ varies for each measurement $m$ as the illuminated spatial position varies for each measurement $m$.
This represents an MMV problem in forward problem formulation.
\subsection{Defect detection and reconstruction}
In our case, defects on a material are physically described as a change of optical absorption coefficient since the defect region (high value: ${\text{a} \sim 0.95}$) differs from the defect-free region (low value: ${\text{a} \sim 0.15}$) in its material properties. Thus, the absorption pattern $a$ (see equation \eqref{eq a}) represents the defect pattern which is of interest. Since we assume blind structured illumination, we can not separate the illumination $I^m_{r,t}$ from the absorption pattern $a$. However, we can reformulate equation \eqref{eq a} by extracting the known illumination pulse time from ${I^m_{r,t} = I_r^m \otimes I_t}$ - with ${I_r^m \in \mathbb{R}^{N_r}}$ and ${I_t \in \mathbb{R}^{N_t}}$ - which has not been modified over the measurements, resulting in:
\begin{equation}
\begin{split}
    T^m &= \Phi_{\text{PSF}} \ast_{r,t} [(I_r^m \otimes I_t) \circ (a \otimes 1)]\\
    &= (\Phi_{\text{PSF}} \ast_t I_t) \ast_r (I_r^m \circ a)\\
    &= \Phi \ast_r x^m_r,
\end{split}
\end{equation}
where $\Phi$ represents the PSF which considers the illumination pulse length.
\subsection{Photothermal super resolution}\label{sec:photothermalSR}
The photothermal super resolution (SR) technique refers to structured illumination in space. This technique promotes high resolvability by e.g. illuminating a specimen using a narrow laser line. The realization of photothermal SR is possible if a lot of measurements/ illuminations are performed with the narrow laser line and is such a MMV problem. In our previous studies \cite{ole}, we have shown that one can scan step by step with a single laser line with submillimeter position shifts to provide high resolvability. This technique relies on spatial frequency mixing of the illumination pattern and the target pattern which is here the absorption pattern. Spatial frequency mixing allows to generate higher frequency components enabling super resolution. Since $I_r^m$ has a certain width in space according to the used narrow laser line width, the use of photothermal SR results in equal spatial distributions for $\sum_m I_r^m$ and $a$ (see Fig. 7 in Ref. \cite{ole}). Hence, the initial goal is to determine the variable $x_r^m$. We measure $T^m$ with the IR camera and we can determine $\Phi$ analytically according to the Green's function approach \cite{psf}. This underlying inverse problem has been considered many times in our studies \cite{qirt,ndte,ole,nsr} and a promising approach to obtain $x_r$ was the block regularization, where we used LASSO variants such as Block Elastic-Net or Block Iterative Shrinkage Thresholding Algorithm (Block ISTA).
\subsection{Block minimization problem}
Before using block regularization we eliminate the time dimension to reduce the data size. The easiest way shown in our previous studies \cite{ndte,ole,nsr} was to extract one thermogram per measurement in the time domain which exhibits the highest SNR - this is the so-called maximum thermogram (MT) method. The MT method has been applied to each measurement $m$ separately so that we can reformulate the description of the underlying reduced (reduc) data by:
\begin{equation} \label{reduced}
    T_{\text{reduc}}^m = \Phi_{\text{reduc}} \ast_r x_r^m
\end{equation}
with ${T_{\text{reduc}}^m \in \mathbb{R}^{N_r}}$ and ${\Phi_{\text{reduc}} \in \mathbb{R}^{N_r}}$.

Using e.g. Block ISTA then tries to minimize the following term: 
\begin{equation} \label{ijosp}
    \min_{\hat{x}_r^m} \sum_{m=1}^{N_{\text{meas}}} \sum_{k=1}^{N_r}\big| \big(\Phi_{\text{reduc}} \ast_r \hat{x}_r^m\big)[k] - T_{\text{reduc}}^m[k]\big|^2 + \lambda \|\hat{x}_r\|_{2,1}
\end{equation}
with the block or joint sparsity term \newline $\|\hat{x}_r\|_{2,1} = \sum_{k=1}^{N_r} \sqrt{\sum_{m=1}^{N_{\text{meas}}} |\hat{x}_r^m[k]|^2}$. Here, the $\ell_{2,1}$-norm is important to take since we are working with blind structured illumination. In our previous studies \cite{ndte,ole,nsr}, we have used the iterative joint sparsity approach where we made use of the Block-ISTA (BISTA) algorithm to minimize the term in \eqref{ijosp} as follows (inspired by Ref. \cite{murray,haltmeier2013block}).
\begin{algorithm} 
    \SetKwInOut{Input}{Input}
    \SetKwInOut{Output}{Output}
    \Input{$\Phi_{\text{reduc}}$, $T^m_{\text{reduc}} $}
    \Output{$\hat{x}^m$}
    $\hat{x}^m_{(0)}=0$\\
    \For{$i = 1 \dots N_{\text{iter}}$}
   {
   $\hat{x}^m_{(i)}=\eta_{\lambda}\left(\hat{x}^m_{(i-1)}-2 \gamma \Phi_{\text{reduc}}\ast_r \left(\Phi_{\text{reduc}}\ast_r \hat{x}^m_{(i-1)} - T^m_{\text{reduc}} \right)\right)$
   }
    \caption{\label{BISTA}Block-ISTA algorithm}
\end{algorithm}

The number of iterations $N_{\text{iter}}$ is chosen such that convergence is reached and the soft block thresholding operator $\eta_{\lambda}$ can be computed by:
\begin{align}
    \eta_{\lambda}(\hat{x}^m)[k]=\max\left\lbrace0,1-\frac{\lambda}{\sqrt{\sum_{m=1}^{N_{\text{meas}}} |\hat{x}^m[k]|^2}}\right\rbrace\hat{x}^m[k]  . \label{blocksoft} 
\end{align}
The step size is defined by ${\gamma\in (0,\frac{1}{L}]}$ where $L$ is the largest Lipschitz constant of the gradient of the data fitting term. The value for $L$ can be determined empirically or by ${L = 2 \pi \|\hat{\Phi}_{\text{reduc}}\|_{\infty}}$ where $\|\cdot\|_{\infty}$ denotes the supremum/ infinity norm and $\hat{\Phi}_{\text{reduc}}$ denotes the Fourier transform of $\Phi_{\text{reduc}}$ \cite{murray}. The value for $\lambda$ is chosen empirically in Algorithm \ref{BISTA} - Block-ISTA, e.g. $\lambda=4e-3$.

\section{Learned Block Iterative Shrinkage Thresholding Algorithm}
To avoid tedious manually selected regularization parameters and to ensure a higher convergence speed than reached by using BISTA, we propose the use of learned block ISTA (LBISTA). LBISTA allows unrolling BISTA as a neural network being able to train the network towards an optimal choice for the regularization parameter layer by layer or even an optimal choice for the weights of $\Phi_{\text{reduc}}$. In the following, we will denote the learning procedure without learning the weights of $\Phi_{\text{reduc}}$, but learning the regularization parameter $\lambda$, as tied learning. If we additionally learn the weights of $\Phi_{\text{reduc}}$ layer by layer we call it untied learning (inspired by Ref. \cite{amp}).

The success of LBISTA is strongly dependent on the choice of the training data and on the implementation of the training (tied or untied learning) which will be explained in detail in the following subsections. 

\subsection{Training data}
To create the training data, we follow the forward problem shown in equation \eqref{reduced}.
To define $x_r^m$, we have different parameters which can be varied: the defect width (see the width of a black stripe in Fig. \ref{specimen}), the laser line width (see the width of a single laser line shown in Fig. \ref{specimen}), the absorption coefficient for a defective and a defect-free region, the number of defects (probability of nonzero in space for $a$ if we assume that the defect-free regions have an absorption coefficient of 0) and the number of laser lines (probability of nonzero in space for $I_r^m$ if we assume that $I_r^m$ equals zero at the non-illuminated positions). Hence, we implemented $x_r^m$ as ${I_r^m \circ a}$, determined $\Phi_{\text{reduc}}$ according to the Green's function approach (see detailed descriptions in Ref. \cite{ole}) and calculated $T_{\text{reduc}}^m$ by a spatial convolution of $x_r^m$ and $\Phi_{\text{reduc}}$ as shown in equation \eqref{reduced}.
\subsection{Training implementation}
Inspired by the code implementation of Ref. \cite{amp}, the training for tied and untied learning is the same, both are based on the use of the Adam optimizer, only the layer definition looks different and depends on a set $V$ with trainable variables and on a number of layers $K$. As we worked with TensorFlow to implement LBISTA, we first defined the layers as follows. Note that we simplified the code representation with $\ast := \ast_r$ and $y^m := T^m_{\text{reduc}}$.
\begin{algorithm}
    \SetKwInOut{Input}{Input}
    \SetKwInOut{Output}{Output}

    \Input{$y^m$}
    \Output{$\hat{x}^m$}
    $\hat{x}^m_{(0)}=B\ast y^m$\\
    $\hat{x}^m_{(1)}=\eta_{\lambda^{(0)}}\left(B\ast y^m\right)$\\
    \For{$i=2,\dots, K$}
   {
   $\hat{x}^m_{(i)}=\eta_{\lambda^{(i-1)}}\left(S\ast \hat{x}^m_{(i-1)}+B\ast y^m \right)$
   }
    \caption{\label{alg:tied learning}LBISTA, layer definition: tied learning}
\end{algorithm}
\begin{algorithm}
    \SetKwInOut{Input}{Input}
    \SetKwInOut{Output}{Output}
    \Input{$y^m$}
    \Output{$\hat{x}^m$}
    $\hat{x}^m_{(0)}=0$\\
    \For{$i=1,\dots, K$}
   {
   $\hat{x}^m_{(i)}=\eta_{\lambda^{(i-1)}}\left(S^{(i-1)}\ast \hat{x}^m_{(i-1)}+B^{(i-1)}\ast y^m \right)$
   }
    \caption{\label{alg: untied learning}LBISTA, layer definition: untied learning}
\end{algorithm}

 We define the set of trainable variables for Algorithm \ref{alg:tied learning}:
\begin{align*}
    V_{tied}^{(0)}&=\emptyset\\
    V_{tied}^{(i)}&=\left\lbrace \lambda^{(i-1)} \right\rbrace,\, i=1,\dots, K \\
    V_{tied} &= \left\lbrace S,B\right\rbrace \cup \bigcup _{i=1}^K V_{tied}^{(i)}.
\end{align*}
We initialize these variables with ${B=2\gamma\Phi_{\text{reduc}}}$, ${S=E-B\ast\Phi_{\text{reduc}}}$, where ${E=[1, 0, \dots, 0]^T\in\mathbb{R}^{N_r}}$ and ${\lambda^{(i)}=4e-3}$ for some step size ${\gamma\in (0,\frac{1}{L}]}$ and all ${i=0,\dots,K-1}$.

For Algorithm \ref{alg: untied learning} we define:
\begin{align*}
    V_{untied}^{(0)}&=\emptyset\\
    V_{untied}^{(i)}&=\left\lbrace S^{(i-1)}, B^{(i-1)},\lambda^{(i-1)} \right\rbrace, i=1,\dots, K \\
    V_{untied}& = \bigcup _{i=1}^K V_{untied}^{(i)}.
\end{align*}
Here we initialize ${B^{(i)}=2\gamma\Phi_{\text{reduc}}}$, ${S^{(i)}=E-B\ast \Phi_{\text{reduc}}}$, for all ${i=0,\dots, K-1}$.
\begin{algorithm}
    \SetKwInOut{Input}{Input}
    \SetKwInOut{Output}{Output}

    \underline{function train} $(t_r, f, x^{*,m},V_{case})$\;
    \Input{Training rate $t_r$, refinements $f$, exact solution $x^{*,m}$ and trainable Variables $V$ with $case=tied$ or $case=untied$}
    \For{$i=0,\dots, K$}
   {
   $loss = \frac{1}{2}\sum_{m=1}^{N_\text{meas}} \|\hat{x}_{(i)}^m-x^{*,m} \|_2^2$\\
   \If{$V_{case}^{(i)}\neq\emptyset$} {
    \While{t $<$ Maxiter}{
    AdamOptimizer($t_r$).min\big($loss$, var list=$V_{case}^{(i)}$\big)\\
    t++\\
    }
    }
    \For{$f_m$ in refinements $f$}
    {
     \While{t $<$ Maxiter}{
        AdamOptimizer($t_r\cdot f_m$).min\big($loss$, var list=$V_{case}$\big)\\
        t++\\
    }
    }
   }

    \caption{LBISTA, implementation of training}
\end{algorithm}

 The training can take a large amount of time depending on the choice of the termination condition and therefore the number of iterations for each layer and each refinement. Hence, the training time also depends on the number of refinements and on the number of layers. The calculation time for one iteration is around $150$\, ms. Within a layer, we usually have 10000 iterations and with each refinement additionally 1000 iterations. We have used the GPU Quadro RTX 8000 to perform the training.

It should be noted that during performing the refinements even $S$ and $B$ are refined in tied case, so that tied learning with refinements kind of resembles untied learning.
\begin{figure*}[!h]
    \centering
    \includegraphics[width = \textwidth]{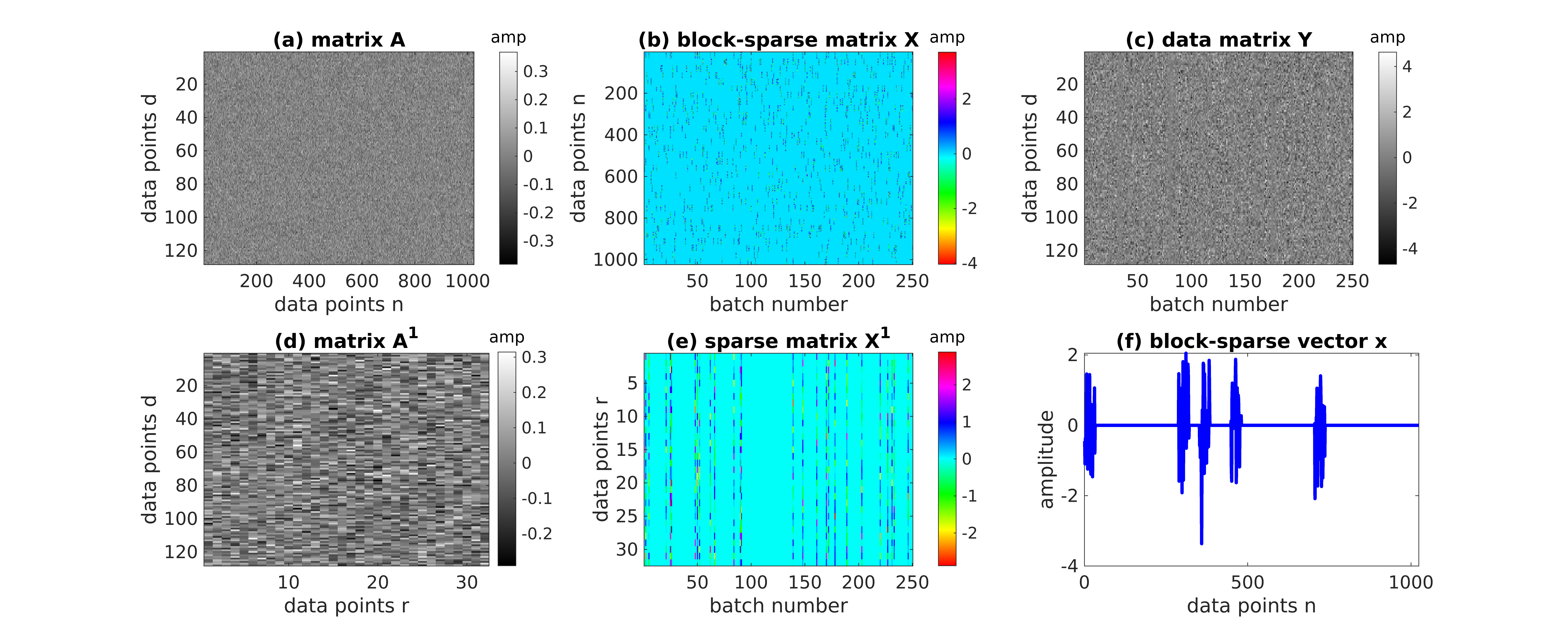}
    \caption{Numerical studies: Datasets used for training - case 1. All shown values and variables are unitless. Further, ${n\in \{1,\,\dots,\,N_r\cdot N_{\text{meas}}\}}$, ${d \in \{1,\,\dots,\,N_d\}}$. (a) A randomly normal Gaussian distributed measurement matrix A, (b) random uniform generated block-sparse matrix $X$, (c): result from the product of (a) and (b), (d): part of (a), whereby matrix $A^1$ is from ${A = [A^1,\,\dots,\,A^{32}]}$ with ${N_{\text{meas}} = 32}$, (e): part of (b) with matrix $X^1$ from ${X = [X^1,\,\dots,\,X^{32}]}$, (f): an exemplary block-sparse vector extracted from (b) for batch number $5$.}
    \label{training_data_case1}
 \vspace*{\floatsep}%
    %\centering
    \includegraphics[width = \textwidth, trim={2cm 0cm 1.5cm 0cm}, clip]{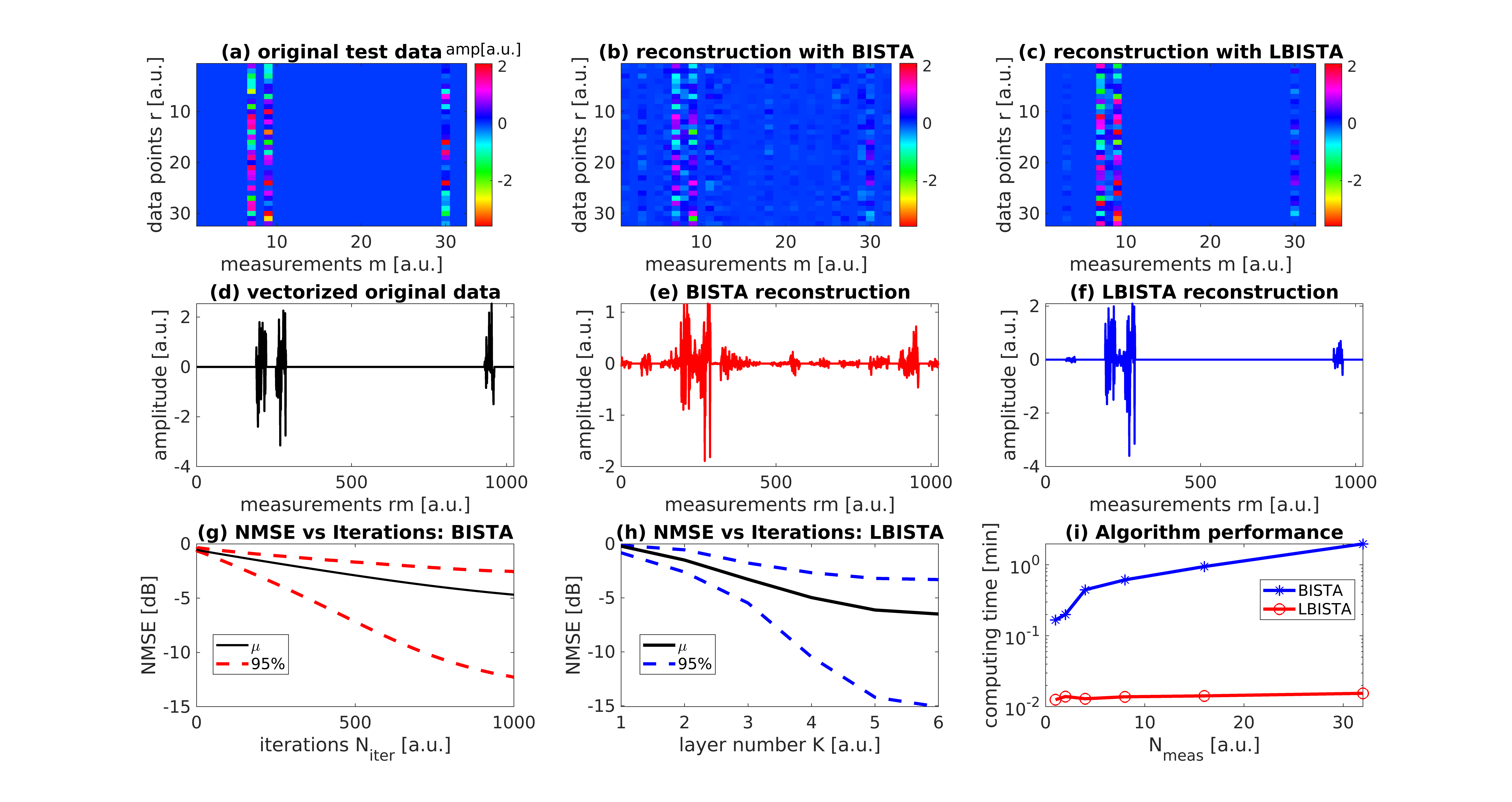}
    \caption{Numerical studies: Reconstruction results using LBISTA vs state-of-the-art BISTA. (a): data to be reconstructed $X_{\text{test,\,case\,1}}$; (b,c): reconstructed data using BISTA or LBISTA by using $Y_{\text{test,\,case\,1}}$ and $A$; (d,e,f): vectorized illustration of (a,b,c); (g,h): normalized mean square errors (NMSE) with $95\%$ confidence interval calculated for reconstructed and original $x$ for 250 different test datasets over number of iterations using BISTA or LBISTA, $\mu$ stands for the mean; (i): performance tests for BISTA with 1000 iterations and LBISTA with 6 layers.}
    \label{test_data_case1}
\end{figure*}

\section{Evaluation of LBISTA}
\subsection{Numerical results}
To study the performance of the LBISTA, we first show numerical studies for the simple product case (case 1) and second, for the convolution case (case 2) which is more related to our examined thermographic problem. More precisely, case 1 investigates the Block-Gaussian product problem ${Y = A \cdot X}$. Case 2 considers the block-sparse convolution problem ${T_{\mathrm{reduc}}^m = \Phi_{\mathrm{reduc}} \ast_r x_r^m}$.

\subsubsection{Case 1: Gaussian measurements}
We use a different designation with $Y$ instead of $T$ since the data here is not relating to the dimensions of the measured data and kept simple and small just for evaluating the LBISTA. More realistic dimensions would lead to a large matrix $A$ which is impossible to process due to the limited RAM of the used GPU. We used the following dimensions for the training data in the evaluation of case 1: ${Y \in \mathbb{R}^{N_d \times N_B}}$, ${A \in \mathbb{R}^{N_d \times N_r \cdot N_{\text{meas}}}}$ and ${X \in \mathbb{R}^{N_r \cdot N_{\text{meas}} \times N_B}}$, whereby $N_B$ denotes the number of training batches and $N_d$ the number of measured data points. According to the block-sparsity definition in Ref. \cite{eldar}, we vectorized the original block-sparse ${N_r \times N_{\text{meas}}}$ dimensional matrices to block-sparse ${N_r \cdot N_{\text{meas}}}$ dimensional vectors. For evaluation we used ${N_r = 32,\,N_{\text{meas}} = 32,\,N_d = 128}$ and ${N_B = 150}$.

The entries $A_{k,l}$ of the matrix $A$ are sampled from a zero-mean Gaussian distribution with standard deviation $\frac{1}{\sqrt{N_d}}$. The column vector ${X_{:,l}\in \mathbb{R}^{N_r \cdot N_{\text{meas}}}}$ is block-sparse, where a certain block is active, i.e., it's $\ell_2$-norm is non-zero, with probability $\text{PNZ}$ (probability of non-zeros). The elements of an active block are sampled from a Gaussian distribution with mean zero and standard deviation of $1$. 
To generate $Y$, Gaussian noise is added. For the determination of Gaussian noise variance $\sigma_{\text{noise}}^2$, we have used: SNR = ${\mu^2/\sigma_{\text{noise}}^2 = 20\,\text{dB}}$ with the squared signal mean ${\mu^2 = \text{PNZ} \cdot \frac{N_r \cdot N_{\text{meas}}}{N_d}}$. Thus, we worked with:
\begin{align}
    Y = A \cdot X.
\end{align}
For better imagination, the training data structure is shown in Fig. \ref{training_data_case1}.

We have used 150 batches to train the network, i.e. 150 different block-sparse vectors.

To evaluate the performance of the LBISTA network, we created a dataset ${Y_{\text{test,\,case\,1}}\in \mathbb{R}^{N_d}}$ based on a certain test data ${X_{\text{test,\,case\,1}} \in \mathbb{R}^{N_r\cdot N_{\text{meas}}}}$ (similarly generated as shown in Fig. \ref{training_data_case1} (b)) which has not been used in training. Thus, we have used: ${Y_{\text{test,\,case\,1}} = A \cdot X_{\text{test,\,case\,1}}}$. For testing, we investigated a compressed sensing case with ${N_r \cdot N_{\text{meas}} = 1024 \gg N_d = 128}$. Hence, the goal is to give the LBISTA network as input $Y_{\text{test,\,case\,1}}$ and $A$ and to obtain as ideal output the $X_{\text{test,\,case\,1}}$ from Fig. \ref{test_data_case1} (a). 

The reconstruction results of $X_{\text{test,\,case\,1}}$ based on LBISTA and state-of-the art BISTA are shown in (b,c). These images have been created on the basis of the number of iterations ${N_{\text{iter}} = 1000}$ for BISTA and number of iterations or layers ${K=6}$ for LBISTA. Vectorizing the images in (a,b,c) results in block-sparse vectors in (d,e,f). Moreover, we evaluated the NMSE for 250 randomly chosen batches that have not been used for training. Hence, the NMSE for the reconstruction using LBISTA and BISTA, examining the error of the estimated data and the original test data, is demonstrated in (g,h), as well as a performance comparison of the LISTA and LBISTA algorithm depending on the number of investigated measurements (f). 

To obtain these results the same initial regularization parameter ${\lambda = 0.1}$ was used. As the BISTA algorithm does not change $\lambda$ over the number of iterations, it cannot reach the same accurate results for a small fixed number of iterations as achieved by using LBISTA that tries to find over the layers an optimal $\lambda$ and therefore exhibits a higher speed of convergence than BISTA. In the 6th layer LBISTA uses $\lambda = 3.3575$ to generate the results shown in Fig. \ref{test_data_case1}. In addition, it should be noted that more accurate results are achieved although less iterations have been used. Comparing Fig. \ref{test_data_case1} (g) with (h) the NMSE curve of BISTA has a much smaller absolute slope than the NMSE curve of LBISTA. The NMSE at layer ${K=6}$ for LBISTA is still ${2\,\text{dB}}$ better than the NMSE at iteration ${N_{\text{iter}} = 1000}$ of BISTA.  

\begin{figure}[h]
    \centering
    \includegraphics[width = 0.45 \textwidth]{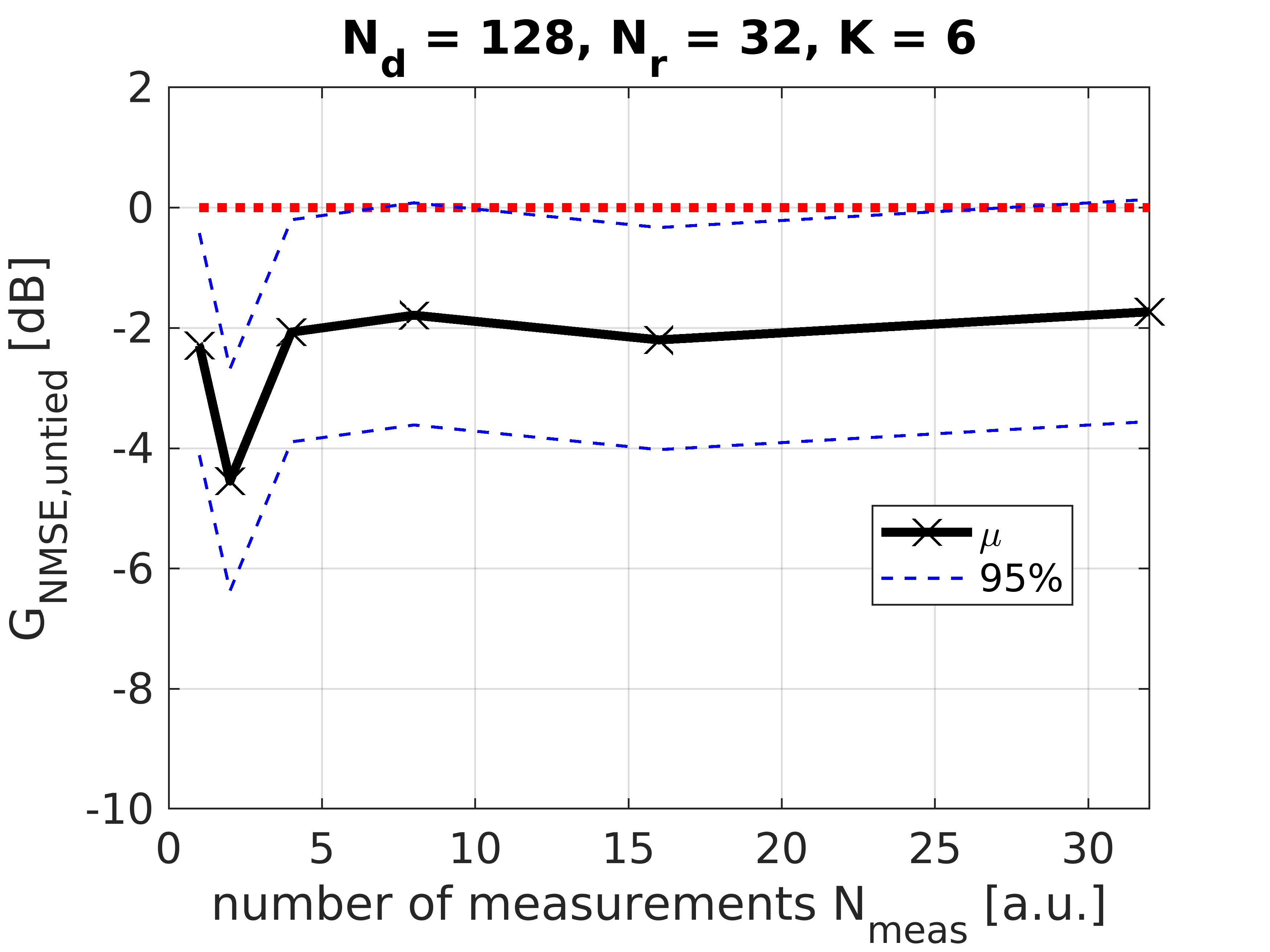}
    \caption{The gain in dB using untied learning instead of tied learning is shown for different measurement numbers. The red dotted line indicates a gain of ${0\,\text{dB}}$ meaning that no improvement has been achieved using untied instead of tied learning. Positive gains indicate a degradation of the NMSE applying untied instead of tied learning.}
    \label{untiedlearningnmsegain}
\end{figure}

The results above for LBISTA are all shown for the tied learning implementation. Using untied learning can further improve the NMSE as additionally the matrices $S$ and $B$ are adjusted layer by layer. Fig.  \ref{untiedlearningnmsegain} shows the NMSE gain induced by applying untied instead of tied learning within the LBISTA network. More precisely, the untied learning NMSE gain $G_{\text{NMSE,\,untied}}$ is calculated by

\begin{align} 
    \begin{split}
    G_{\text{NMSE,\,untied}} &= \text{NMSE}_{\text{untied}}-\text{NMSE}_{\text{tied}} \\
    \text{NMSE} &= 10 \cdot \text{log}_{10}\bigg(\frac{\|x^m-\hat{x}^m\|_2^2}{\|x^m\|_2^2}\bigg).
    \end{split}
\end{align}

Fig. \ref{untiedlearningnmsegain} shows that the gain does not really increase or decrease with larger datasets. There is a constant improvement applying untied learning instead of tied learning. The ${95\,\%}$ confidence interval analysis has been performed based on $250$ different test datasets. Very rarely, the application of untied learning does not lead to an improvement of the NMSE (see upper bound given by ${95\,\%}$-analysis).

\subsubsection{Case 2: Convolution measurements}
The convolution case is more related to our investigated use case in photothermal super resolution imaging. Therefore, we use the same notation as shown in eq. \eqref{reduced} with ${y_{\text{train}}^m = \Phi_{\text{reduc}} \ast_r x_{\text{train}}^m}$. An analytical formulation of $\Phi_{\text{reduc}}$ can be found in Ref. \cite{ndte}. Table \ref{tab:training_data_case2} shows the used parameters for training with synthetic data based on ${x_{\text{train}}^m = I_{\text{train}}^m \circ a}$. More precisely, the cells corresponding to defect pattern determine the values in $a$ and the cells corresponding to illumination pattern the values in ${I_{\text{train}}^m \in \mathbb{R}^{N_r}}$. Further, Gaussian noise is added to generate $y_{\text{train}}^m$ based on the shown SNR value in the table, similarly calculated as explained in case 1. The LBISTA network is trained with tied and untied learning for the dimensions ${N_r = 1280}$ (number of pixels), ${N_{\text{meas}} = 150}$ (number of measurements based on differently chosen illumination and therefore differently generated $x^m$). 

\begin{table}[h]
    \centering
    \begin{tabular}{|c|c|}
    \hline
         \textbf{Gaussian Noise in $y_{\text{train}}^m$}& SNR = 8 dB \\ \hline
         \textbf{Defect pattern} & defect width = $1\,$mm \\ \hline
          & defect sparsity (PNZ) = $0.01$
          \\ \hline
          & absorption coefficient = $\{0,\,1\}$ \\ \hline 
          \textbf{Illumination pattern} & laser line width = $0.8\,$mm \\ \hline
           & illumination sparsity (PNZ) = $0.01$ \\ \hline
           \textbf{Training parameters} & refinements $f$: $f_m = \{0.5,\,0.1,\,0.05\}$ \\ \hline
            & training rate $t_r = 0.001$ \\ \hline
             & initial lambda $\lambda = 0.004$ \\ \hline
             & number of layers $K = 6$ \\ \hline
             & step size $\gamma = \frac{1}{\sqrt{2}}$ \\ \hline
             & batch number $N_B = 150$ \\ \hline
             tied learning& max. number of iterations $Maxiter = 10^5$ \\ \hline
             untied learning& max. number of iterations $Maxiter = 10^4$ \\ \hline

    \end{tabular}
    \caption{Used parameters in LBISTA network for problem definition (defect and illumination pattern) and for training in the convolution case.}
    \label{tab:training_data_case2}
\end{table}
The maximum number of iterations $Maxiter$ has been decreased for untied learning since untied learning updates more variables causing higher computation times for each iteration. 
\begin{figure*}[!htbp]
    \centering
    \includegraphics[width = 1 \textwidth,trim = {3cm 0.5cm 3cm 0cm},clip]{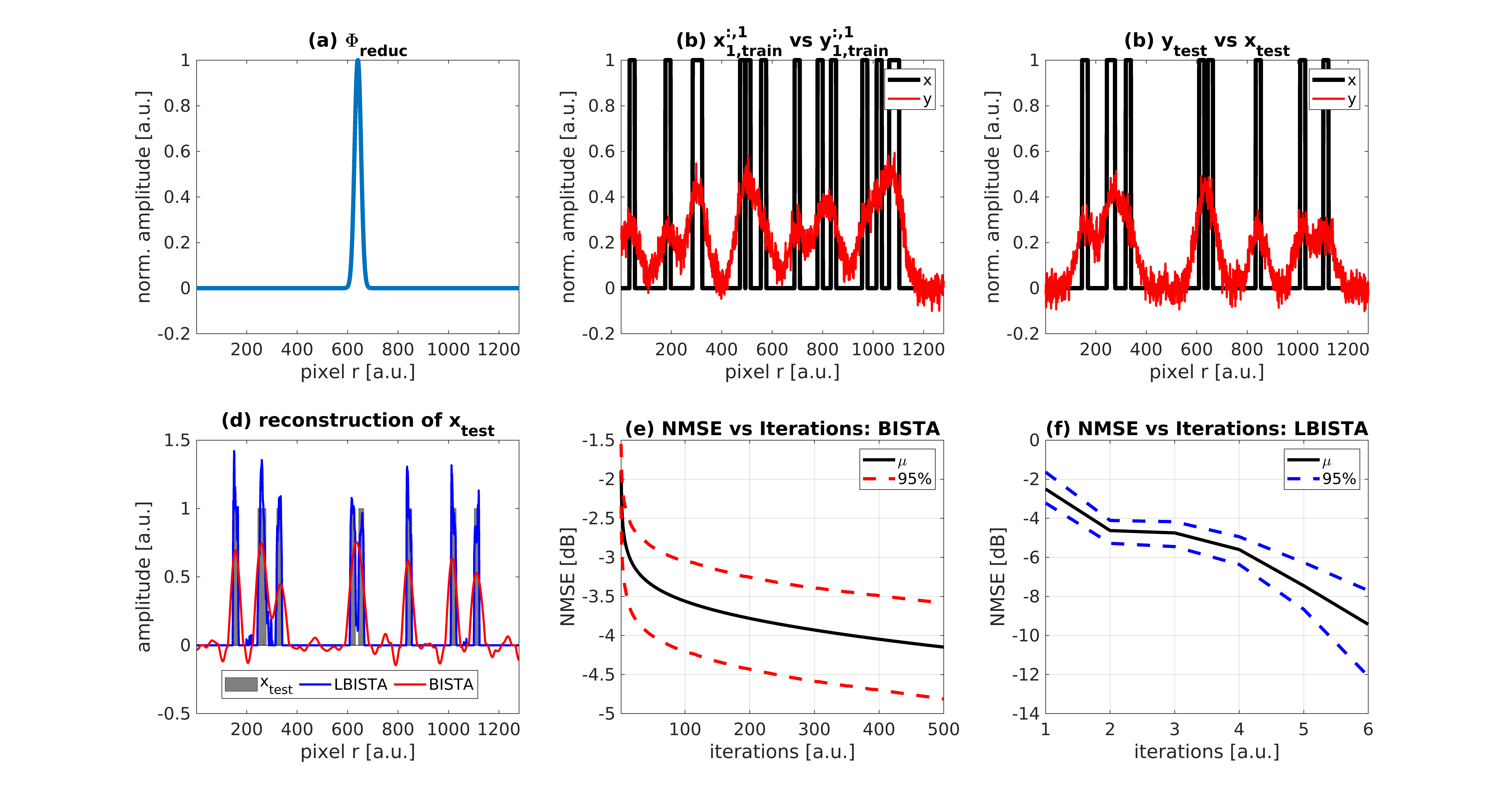}
    \caption{Exemplary synthetic training and synthetic test data for case 2 (convolution) as well as NMSE performance studies. (a): shape of used $\Phi_{\text{reduc}}$ in training, (b): comparison of the curve shapes for exemplary training datasets with e.g. ${x_{1,\,\text{train}}^{:,1} \in \mathbb{R}^{N_r}}$, (c): comparison of the curve shapes for exemplary test datasets, (d): reconstruction of $x_{\text{test}}$ using convolution-based tied LBISTA network ($6$ iterations/layers) vs BISTA ($1000$ iterations), (e): NMSE over iterations for BISTA with $95\,\%$ confidence interval, (f): NMSE over iterations for tied LBISTA with $95\,\%$ confidence interval}
    \label{fig:training_data_case2}
\end{figure*}
The $\Phi_{\text{reduc}}$ used in training will also be used in the next section to evaluate real experimental data. In the following, the training data in case 2 is determined with ${x^{n,p}_{\text{train}}, y^{n,p}_{\text{train}}\in \mathbb{R}^{N_r}}$ and ${n \in \{1,\,\dots,\,N_B\}}$, ${p \in \{1,\,\dots,\,N_{\text{meas}}\}}$, so that\\ $\left(X_{i\text{, train}}\right)_{i=1}^{N_B}=\left(\begin{bmatrix}
x^{1,1}_{i\text{, train}} & \dots & x^{1,N_{\text{meas}}}_{i\text{, train}} \\ 
\dots & \dots & \dots \\ 
x^{N_r,1}_{i\text{, train}} & \dots & x^{N_r,N_{\text{meas}}}_{i\text{, train}} \\ 
\end{bmatrix}\right)_{i=1}^{N_B} \in \mathbb{R}^{N_B \times N_r \times N_{\text{meas}}}$ and\\ $\left(Y_{i\text{, train}}\right)_{i=1}^{N_B} = \left(\begin{bmatrix}
y^{1,1}_{i\text{, train}} & \dots & y^{1,N_{\text{meas}}}_{i\text{, train}} \\ 
\dots & \dots & \dots \\ 
y^{N_r,1}_{i\text{, train}} & \dots & y^{N_r,N_{\text{meas}}}_{i\text{, train}} \\ 
\end{bmatrix}\right)_{i=1}^{N_B} \in \mathbb{R}^{N_B \times N_r \times N_{\text{meas}}}$. Exemplary training data is shown visually in Fig. \ref{fig:training_data_case2} (b).

Looking at Fig. \ref{fig:training_data_case2} (e,f), it is obvious that LBISTA reaches a smaller normalized mean square error (NMSE) much faster than BISTA. According to the curves in (d), $x_{\text{test}}$ LBISTA outperforms BISTA in terms of reconstruction using $\Phi_{\text{reduc}}$ (see (a)) and $y_{\text{test}}$ (see (b)). Of course, it strongly depends on the choice of $\lambda$ how well BISTA reconstructs. Therefore, we chose the same initial values of $\lambda$ for BISTA and LBISTA to have a better comparison.

Thus, Fig. \ref{fig:training_data_case2} shows the results with synthetic test data $y_{\text{test}}$ and ground truth $x_{\text{test}}$ which are very similar to the datasets used for training $Y_{\text{train}},\,X_{\text{train}}$.
In the following section we use $T_{\text{reduc}}$ as real measurement test data instead of $y_{\text{test}}$, but still the same training as shown in case 2 based on synthetic data.

\subsection{Evaluation of LBISTA with experimental data from active thermal imaging}
Instead of creating test data synthetically as in the previous section, we apply the LBISTA based neural network to real experimental data.
The experimental data was measured with the IR camera (InfraTec ImageIR 9300, ${1280 \times 1024}$ pixels full frame, spectral range: ${3-5\,\mu\text{m}}$). We have used the same data set as in a previous publication (see data in Ref. \cite{ndte} based on the MT method) to have a reference. We have used the IR camera in transmission configuration (illumination on the front of the specimen and observation of the back side with the IR camera). The specimen has a thickness of $3\,$mm and 5 blackened stripe pairs on the front side as shown in Fig. \ref{specimen}. All in all, we performed around 150 measurements, resulting in ${m = {1 \dots 150}}$, 30 measurements per blackened stripe pair. Within these 30 measurements, we shifted the position by ${0.4\,\text{mm}}$ twice: once after the first ten measurements and again after the second ten measurements. After 30 measurements the position was shifted to the next blackened stripe pair and the process is repeated until we scanned the whole specimen (see Fig. \ref{specimen} for better understanding). Each measurement differs by randomly (uniform distribution) switching on a certain number of laser lines out of a maximum of twelve laser lines (at least one laser line is turned on for each measurement).
\begin{figure}[!h]
    \centering
    \includegraphics[width = 0.5 \textwidth]{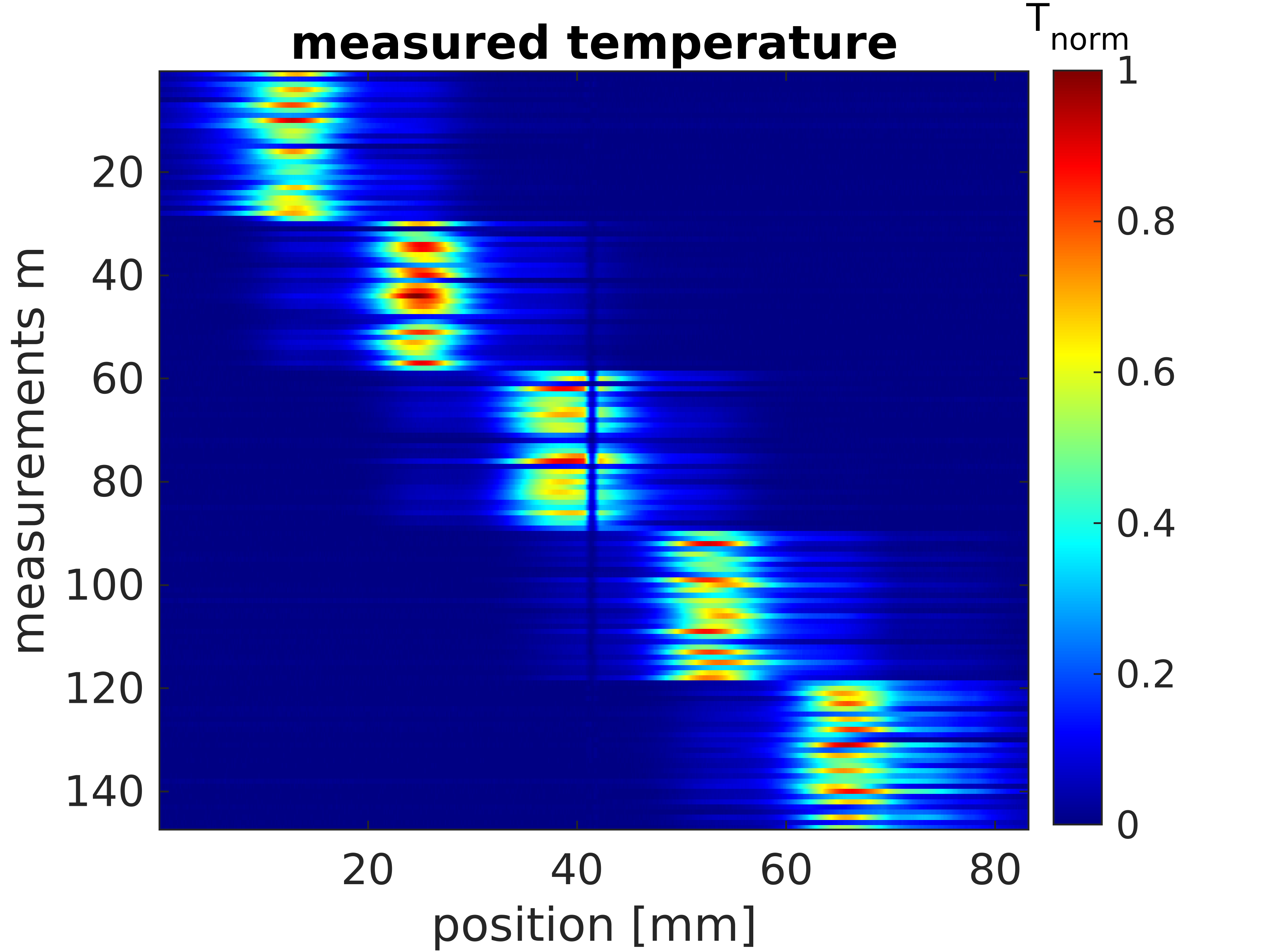}
    \caption{Test data from photothermal structured illumination measurements. To obtain this image, averaging over the vertically arranged pixels as well as applying the maximum thermogram method (see Ref. \cite{ndte,ole,nsr}) to eliminate the time dimension is necessary. At position = ${41\,\text{mm}}$ a blue marker line can be observed which refers to the marker line on the investigated surface of the specimen (see Fig. \ref{erklaerbild}).}
    \label{experimentaltestdata}
\end{figure}
\begin{figure}[!h]
    \centering
    \includegraphics[width = 0.5 \textwidth]{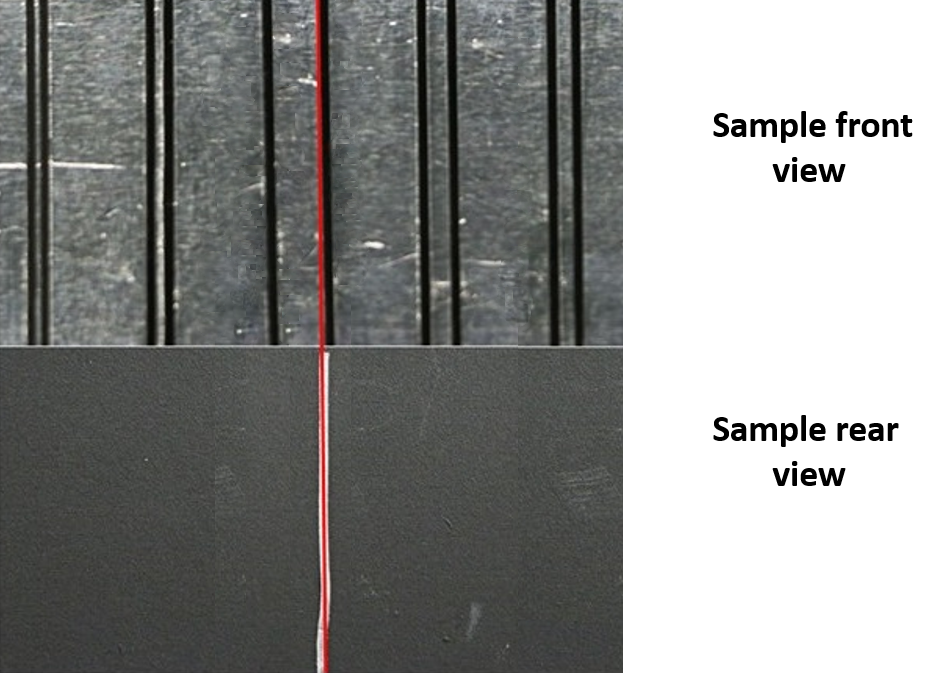}
    \caption{Shown sample front and rear view to understand origin of white marker line at position ${41\,\text{mm}}$ from Fig. \ref{experimentaltestdata}. The rear side of the sample has been observed by the IR camera and the front side has been excited by a laser array. The defects on the front side are shown by blackened stripes (5 defect pairs). The marker line on the rear side is on the same position as the left edge of the right stripe from the third stripe pair.
    The distances between the stripes within the blackened stripe pairs are from left to right: $0.5$, $1$, $3$, $2$, $1.3\,$mm. A stripe is ${1\,\text{mm}}$ wide and the distance of one stripe pair to another is around ${10\,\text{mm}}$.}
    \label{erklaerbild}
\end{figure}
\begin{figure*}[!htbp]
    \centering
    \includegraphics[width =\textwidth]{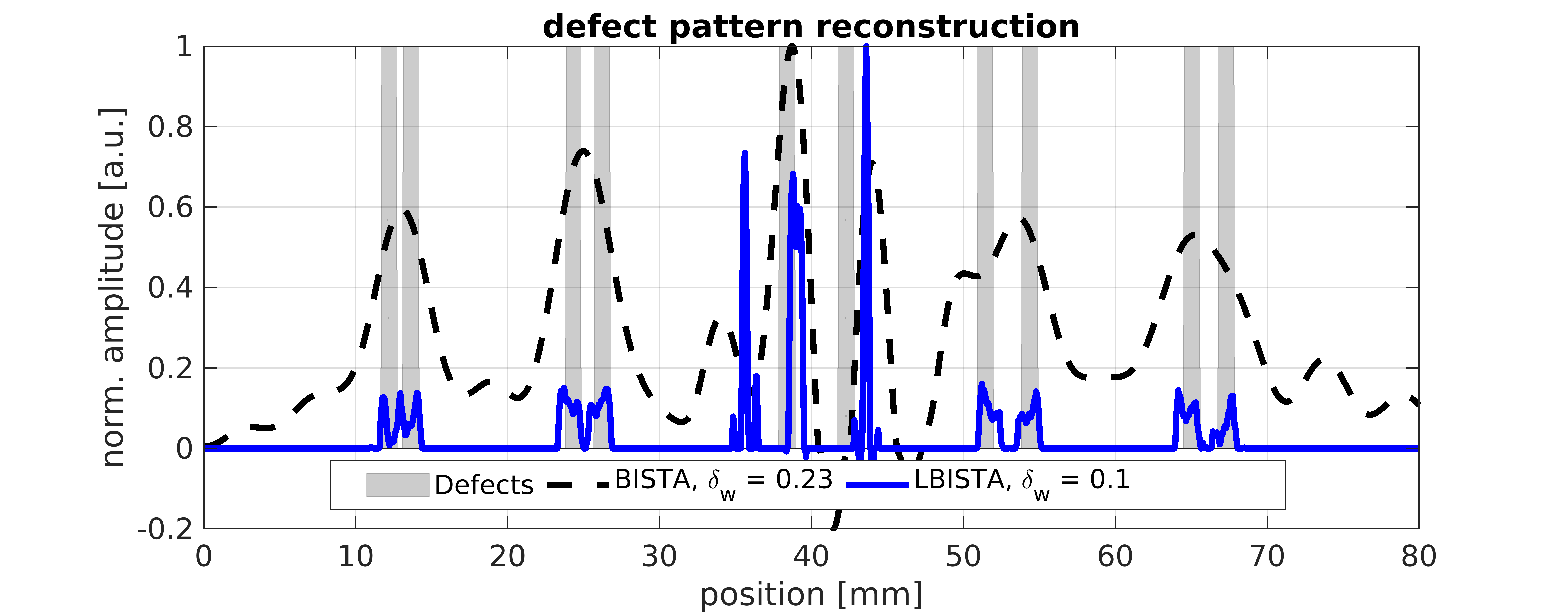}
    \caption{Pattern reconstruction of blackened stripes as shown in Fig. \ref{erklaerbild} using BISTA and tied LBISTA. The parameters to obtain the LBISTA result are shown in table \ref{tab:training_data_case2}. Hence, synthetic training data has been used. The BISTA result has been created by using ${\gamma = \frac{1}{\sqrt{2}}}$, ${\lambda = 0.004}$, ${N_{\text{iter}} = 500}$ (the same values have been chosen in Ref. \cite{ndte} where Block-FISTA has been used instead of BISTA). The marker line can be seen in the BISTA curve exhibiting normalized amplitude values $< 0$ at position ${41\,\text{mm}}$.}
    \label{fig:firstresults}
    \end{figure*}
    \begin{figure*}[!hbtp]
  %  \end{minipage*}
    \vspace*{\floatsep}%
    \includegraphics[width = \textwidth,trim={1.5cm 0 2cm 0},clip]{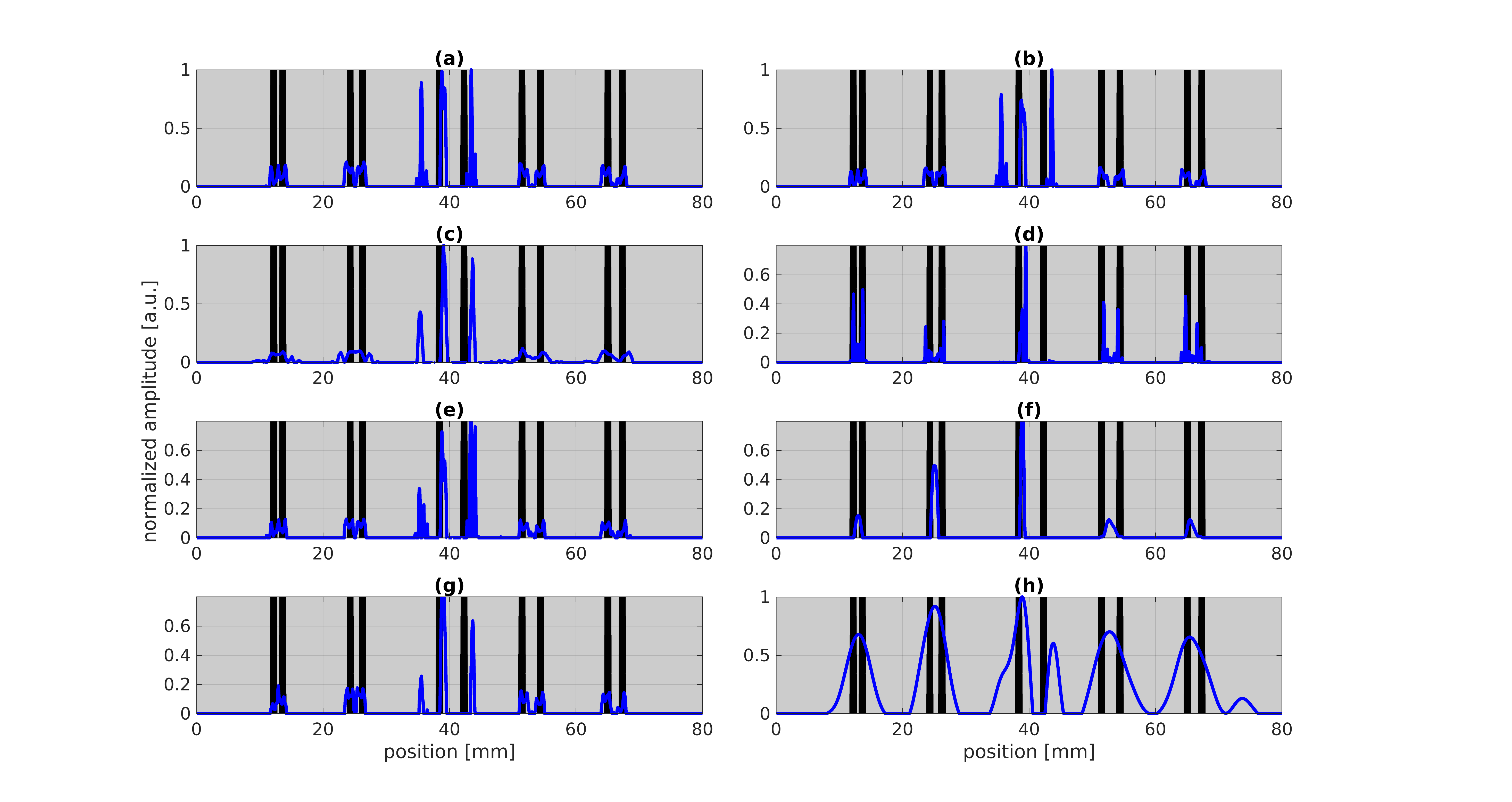}
    \caption{Parameter studies for tied learning LBISTA network applied to experimental test data shown in Fig.\ref{experimentaltestdata}. The blackened areas represent the blackened stripes for better orientation. The blue curves stand for the result after applying the tied LBISTA network to the experimental test data. All images are based on the parameter choice shown in table \ref{tab:training_data_case2}. The following parameters have been changed, respectively: (a) ${N_B = 100}$, (b) ${N_B = 200}$, (c) ${\text{defect width} = 0.5\,\text{mm}}$, (d) ${\text{defect width} = {2\,\text{mm}}}$, (e) defect sparsity ${\text{PNZ} = 0.005}$, (f) defect sparsity ${\text{PNZ} = 0.03}$, (g) only one refinement ${f_m=\{0.5\}}$, (h) only three layers ${K=3}$ and one refinement ${f_m=\{0.01\}}$.}
    \label{fig:parameterstudies_tied}
\end{figure*}
\begin{figure*}[!htbp]
    \centering
    \includegraphics[width = \textwidth,trim={1.5cm 0 2cm 0},clip]{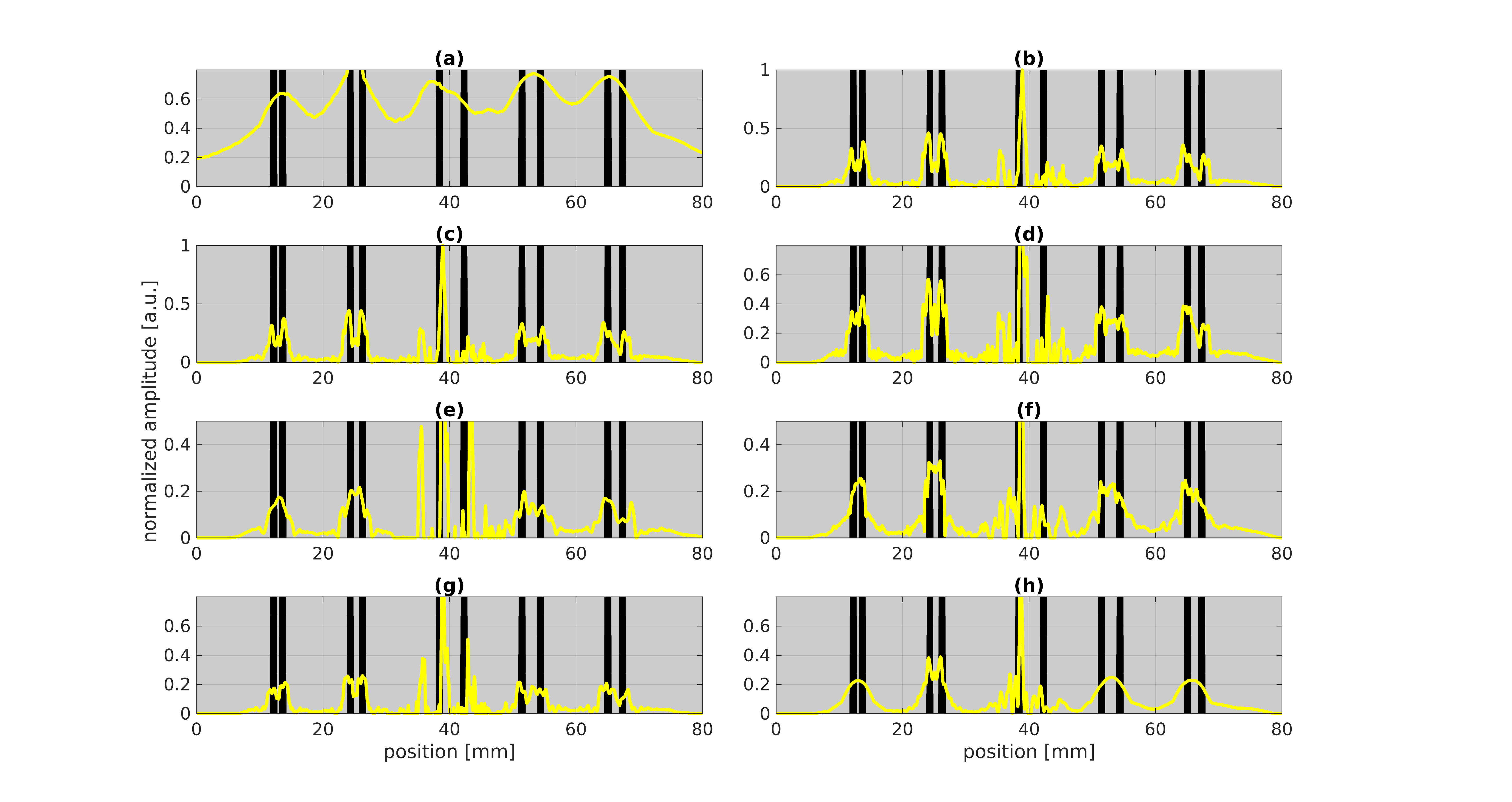}
    \caption{\label{fig:parameterstudies_untied}Parameter studies for untied learning LBISTA network applied to experimental test data shown in Fig.\ref{experimentaltestdata}. The blackened areas represent the blackened stripes for better orientation. The yellow curves stand for the result after applying the untied LBISTA network to the experimental test data. All images are based on the parameter choice shown in table \ref{tab:training_data_case2}. The following parameters have been changed, respectively: (a) ${\text{absorption coefficient} = \{0.3,\,0.7\}}$, (b) ${Maxiter = 10^5}$, (c) ${N_B = 100}$, (d) ${N_B = 200}$, (e) ${\text{defect width} = 0.5\,\text{mm}}$, (f) ${\text{defect width} = 2\,\text{mm}}$, (g) defect sparsity ${\text{PNZ} = 0.005}$, (h) defect sparsity ${\text{PNZ} = 0.03}$.}
\end{figure*}
Thus, the experimental testing data - after removing the dimension $y$ by calculating the mean over the vertically arranged pixels and eliminating the time dimension by applying the maximum thermogram method (Ref. \cite{ndte,ole,nsr}) - looks like in Fig. \ref{experimentaltestdata}.

The figure clearly shows the 30 measurements for each blackened stripe pair. Nevertheless, there is a disturbance in the data at position ${41\,\text{mm}}$ represented by a straight line over all measurements. This disturbance comes from the fact that the specimen that has been investigated had a marker line on the rear side to have an orientation about the defects in thermographic transmission configuration. For better understanding, Fig. \ref{erklaerbild} shows this marker line as well as the investigated defect pattern of the specimen.

In the following, we compare the result after applying BISTA or LBISTA by using as input $T^m$ (see Fig. \ref{experimentaltestdata}) and $\Phi_{\text{reduc}}$ (see Fig. \ref{fig:training_data_case2} (a)). Calculating the sum over all measurements as explained in section \ref{sec:photothermalSR} results in an approximation of the defect pattern. These results are shown in Fig. \ref{fig:firstresults}.
The dashed curve indicates the outcome of the BISTA algorithm and shows only more or less good indications for the third blackened stripe pair with the largest distance between the stripes (around position ${37 \dots 44\,\text{mm}}$). In contrast, the application of LBISTA results in very good indications for three of five stripe pairs. Only the stripes with the largest distance to each other and the stripes with the closest distance to each other (position ${11 \dots 14\,\text{mm}}$) cannot be clearly recognized. The other three stripes are very well reconstructed, even the stripe width can be recognized. Since the marker line at position $41\,$mm disturbs the pattern recognition, the stripes with the largest distance to each other are hard to resolve.
For quantitative comparison of the BISTA and LBISTA result, we used the 1-Wasserstein distance $\delta_w$ \cite{clement2008elementary} with ${\delta_w^{\text{max}} = 1}$ and ${\delta_w^{\text{min}} = 0}$. 

As the parameters from table \ref{tab:training_data_case2} have been used to train the LBISTA network to generate these results, it is still unclear whether these parameters are the best choice. Therefore, we varied some of the parameters to study their impact on the result. Further, we investigated the effect of using untied learning instead of tied learning. All the figures for the parameter studies are shown in Fig. \ref{fig:parameterstudies_tied} for tied learning and in Fig. \ref{fig:parameterstudies_untied} for untied learning. 

Fig.\ref{fig:parameterstudies_tied}:
\begin{itemize}
    \item Training batch size (a,b): Varying the number of batches between ${N_B = 100}$ and ${N_B = 200}$ does not really change the result.
    \item Defect pattern (c,d,e,f): In contrast, changing the sparsity of the defects from ${\text{PNZ} \in [0.005,\,0.03]}$ (c.f. (e,f)) or the defect width (see (c,d)) leads to significant changes in the results such that the result in (d) is even able to clearly detect both stripes in the first stripe pair with the smallest distance between the stripes. In (f) obviously a too small sparsity have been used as the probability of non-zero is quite high in comparison of the chosen PNZ in (e). Thus, it is obviously beneficial to know roughly the sparsity.
    \item Training rate and number of layers (g,h): The results shown in (g,h) refer to variations of the training rate and the number of layers. In (g) we get similarly good results just by using one refinement instead of three refinements as used in Fig. \ref{fig:firstresults}. Only using one refinement could save a lot of time during training. In (h), a rather bad result is shown, where we used one refinement with a very small training rate and only three layers. This rather bad result is caused by using too few layers and a too small training rate. These parameters should be chosen high enough so that the LBISTA network reaches convergence within training.
\end{itemize}

Fig. \ref{fig:parameterstudies_untied}:
\begin{itemize}
    \item Absorption coefficient and training iterations (a,b): In contrast to tied learning, untied learning exhibits significant changes by varying the absorption coefficient in training. According to our studies, an absorption coefficient of ${\{0,\,1\}}$ is a good choice. Further, increasing the maximum number of iterations as shown in (b) can enhance the reconstruction quality so that all stripes could be indicated very well except for the middle stripe pair (most likely due to the marker line).
    \item Training batch size (c,d): The result in (c) using less batches and iterations as in (b) shows that similarly good results can be achieved. Increasing the number of batches to ${N_B = 200}$ as shown in (d) can further enhance the reconstruction quality as now even the middle defect pair could be clearly resolved. 
    \item Defect width (e,f): The variation of the defect width rather degrades the reconstruction result as shown in (e,f). This means that the default choice of the defect width as shown in table \ref{tab:training_data_case2} performs best for untied learning. However, with tied learning (see Fig. \ref{fig:parameterstudies_tied} (c,d)) we could see improvements changing the parameter of the defect width. We could imagine that the reason is based on learning $\Phi_{\text{reduc}}$ in the untied learning algorithm. 
    \item Defect sparsity (g,h): The variation of the sparsity in untied learning in (g,h) confirms our investigations in  tied learning (see (e,f)) as the result deteriorates using a too small sparsity of $\text{PNZ} = 0.03$.
\end{itemize}
\section{Conclusion and outlook}
The application of the LBISTA based neural network leads to much more promising results than the state-of-the-art BISTA. With LBISTA, we can omit manually selected regularization parameters, we obtain the results faster and with higher reconstruction qualities/ smaller NMSE for a small fixed number of iterations compared to BISTA. This paper showcases the improvements in reconstruction for three different cases: 
\begin{enumerate}
    \item Generic Gaussian measurements (see Fig. \ref{test_data_case1})
    \item Convolution measurements with synthetic test data (see Fig. \ref{fig:training_data_case2})
    \item Convolution measurements with experimental test data from photothermal measurements (see outstanding results for tied learning in Fig. \ref{fig:firstresults} and Fig. \ref{fig:parameterstudies_tied} (d) and for untied learning e.g. in Fig. \ref{fig:parameterstudies_untied} (d))
\end{enumerate}
Thereby 2. and 3. are based on the same training data set. In all cases, it could be observed that untied learning provides more reliable results than tied learning in terms of reconstruction quality. Moreover, the parameter studies encourages to use as many batches, iterations and layers as possible to achieve high reconstruction qualities. Further, a very precise model of the experiment is necessary to provide accurate reconstruction results.

Thus, the application of learned regularization algorithms, such as the proposed LBISTA is highly recommended and attractive for industrial applications where the user does not have to choose regularization parameters manually and benefits from the remarkable speed of convergence of LBISTA. LBISTA therefore enables a less complicated evaluation of the photothermal super resolution data and offers the possibility of reliable in-situ inspections.

As an outlook, we will study other learned block regularization techniques based on unfolding algorithms such as FISTA or Elastic-Net. In addition, we will further study how to increase the performance of the training to train with larger data sets where we do not have to eliminate the time dimension or to calculate the mean over the vertically arranged pixels.
\section*{Acknowledgment}
We would like to thank Osman Musa from TU Berlin for discussing the studies to and assisting in the implementation of LBISTA. Further, we are very grateful to work on experimental results with the specimen provided by Günther Mayr and his team from FH Oberösterreich. \newline
Gefördert durch die Deutsche Forschungsgemeinschaft (DFG) - 400857558, funded by the Deutsche Forschungsgemeinschaft (DFG, German Research Foundation) - 400857558.

%\section*{Acknowledgement}
%We acknowledge help with the production of the spot welds and with the chisel test by Hubert Suwala.

\addtolength{\textheight}{-10cm} 

%%%%%%%%%%%%%%%%%%%%%%%%%%%%%%%%%%%%%%%%%%%%%%%%%%%%%%%%%%%%%%%%%%%%%%%%%%%%%%%%

%%%%%%%%%%%%%%%%%%%%%%%%%%%%%%%%%%%%%%%%%%%%%%%%%%%%%%%%%%%%%%%%%%%%%%%%%%%%%%%%

%%%%%%%%%%%%%%%%%%%%%%%%%%%%%%%%%%%%%%%%%%%%%%%%%%%%%%%%%%%%%%%%%%%%%%%%%%%%%%%%

%%%%%%%%%%%%%%%%%%%%%%%%%%%%%%%%%%%%%%%%%%%%%%%%%%%%%%%%%%%%%%%%%%%%%%%%%%%%%%%%

%\bibliographystyle{IEEEtran}

%\bibliography{ref}

\begin{thebibliography}{99}

\bibitem{cs} DL Donoho et al., ``Compressed Sensing,'' in \emph{IEEE Transactions on Information Theory}, vol. 52, no. 4, 2006.

\bibitem{cs2} Richard G. Baraniuk \emph{IEEE Signal Processing Magazine}, vol. 24, no. 4, 2007.

\bibitem{fista} Amir Beck and Marc Teboulle, ``A fast iterative shrinkage-thresholding algorithm for linear inverse problems'' \emph{SIAM journal on imaging sciences}, vol. 2, no. 1, pp. 183--202, 2009.

\bibitem{lista} Karol Gregor and Yann LeCun, ``Learning Fast Approximations of Sparse Coding,'' \emph{Proceedings of the 27th International Conference on Machine Learning}., pp. 399--406, 2010.

\bibitem{alista} Jialin Liu, Xiahoan Chen, Zhangyang Wang and Wotao Yin, ``ALISTA: Analytic Weights Are As Good As Learned Weights in LISTA,'' \emph{International Conference on Learning Representations}., 2019.

\bibitem{lista-at} Dohyun Kim and Daeyoung Park, ``Element-Wise Adaptive Thresholds for Learned Iterative Shrinkage Thresholding Algorithms,'' \emph{IEEE Access}, vol.8, pp. 45874--45886, 2020.

\bibitem{murray} Todd W. Murray et al., ``Super-resolution photoacoustic microscopy using blind structured illumination,'' \emph{Optica}., vol.4, no.1, 2017.

\bibitem{APL} Peter Burgholzer et al., ``Super-resolution thermographic imaging using blind structured illumination,'' \emph{Applied Physics Letters}., vol.111, no.3, 2017.

\bibitem{qirt} Peter Burgholzer et al., ``Blind structured illumination as excitation for super-resolution photothermal radiometry,'' \emph{Quantitative InfraRed Thermography Journal}., pp. 1--11, 2019.

\bibitem{ndte} Samim Ahmadi et al., ``Photothermal super resolution imaging: A comparison of different thermographic reconstruction techniques,'' \emph{NDT \& E International Journal}., vol. 11 , 2020.

\bibitem{ole} Samim Ahmadi et al., ``Super resolution laser line scanning thermography,'' \emph{Optics and Lasers in Engineering}., vol. 134, 2020.

\bibitem{nsr} Samim Ahmadi et al., ``Laser excited super resolution thermal imaging for nondestructive inspection of internal defects,'' \emph{arXiv}., eprint: 2007.03341, 2020.

\bibitem{inverse_prob} Kurt Bryan and Lester F Caudill Jr., ``An inverse problem in thermal imaging,'' \emph{SIAM journal on Applied Mathematics}., vol.56, no.3, pp. 715--735, 1996.

\bibitem{bayes} Jingbo Wang and Nicholas Zabaras, ``Hiearchical Bayesian models for inverse problems in heat conduction,'' \emph{Inverse Problems}., vol.21, no.1, 2004.

\bibitem{group_lasso} X. Lv, G. Bi and C. Wan, ``The Group Lasso for Stable Recovery of Block-Sparse Signal Representations,'' \emph{IEEE Transactions on Signal Processing}., vol.59, no.4, 2011.

\bibitem{eldar} Eldar, Yonina C and Bolcskei, Helmut, ``Block-sparsity: Coherence and efficient recovery,''\emph{2009 IEEE International Conference on Acoustics, Speech and Signal Processing}, 2009.

\bibitem{psf} Kevin Cole et al., ``Heat conduction using Greens functions,'' \emph{Taylor \& Francis}., 2010.

\bibitem{haltmeier2013block}  
Markus Haltmeier, ``Block-sparse analysis regularization of ill-posed problems via L2,1-minimization,'' \emph{18th International Conference on Methods \& Models in Automation \& Robotics (MMAR)}., 2013.

\bibitem{amp} M. Borgerding, P. Schniter and S. Rangan, ``AMP-Inspired Deep Networks for Sparse Linear Inverse Problems,'' \emph{IEEE Transactions on Signal Processing}., vol. 65, no. 16, pp. 4293--4308 2017.

\bibitem{clement2008elementary} Clement, Philippe and Desch, Wolfgang, ``An elementary proof of the triangle inequality for the Wasserstein metric,''
\emph{Proceedings of the American Mathematical Society}., vol.136, no.1, 2008.

\end{thebibliography}

\end{document}